\documentclass[letterpaper]{article} 
\usepackage{aaai2026}  
\usepackage{times}  
\usepackage{helvet}  
\usepackage{courier}  
\usepackage[hyphens]{url}  
\usepackage{graphicx} 
\usepackage{multicol}
\usepackage{multirow}
\usepackage{booktabs} 
\usepackage{todonotes}
\usepackage{amsmath}
\usepackage{amssymb}
\usepackage{subfigure}

\usepackage{natbib}  
\usepackage{caption} 
\usepackage{algorithm}
\usepackage{algorithmic}
\newcommand{\modelname}{L-HAKT}
\usepackage{newfloat}
\usepackage{listings}
\DeclareCaptionStyle{ruled}{labelfont=normalfont,labelsep=colon,strut=off} 
\floatstyle{ruled}
\newfloat{listing}{tb}{lst}{}
\floatname{listing}{Listing}
\title{Towards LLM-Empowered Knowledge Tracing via LLM-Student Hierarchical Behavior Alignment in Hyperbolic Space}
\author{
    Xingcheng Fu\textsuperscript{\rm 1,2}, 
    Shengpeng Wang\textsuperscript{\rm 1,2},
    Yisen Gao\textsuperscript{\rm 2,3}, 
    Xianxian Li\textsuperscript{\rm 1,2},
    Chunpei Li\textsuperscript{\rm 1,2},
    Qingyun Sun\textsuperscript{\rm 2,4},
    Dongran Yu\textsuperscript{\rm 1,2}
}
\affiliations{
    \textsuperscript{\rm 1}Key Lab of Education Blockchain and Intelligent Technology, Guangxi Normal University, Guilin, China\\
    \textsuperscript{\rm 2}Guangxi Key Lab of Multi-source Information Mining \& Security, Guangxi Normal University, Guilin, China\\
    \textsuperscript{\rm 3} Computer Science and Engineering
    , The Hong Kong University of Science and Technology, Hong Kong, China\\
    \textsuperscript{\rm 4}School of Computer Science and Engineering, Beihang University, Beijing, China\\

    \{fuxc, lixx, yudran, SPalmW\}@gxnu.edu.cn, sunqy@buaa.edu.cn
}

\usepackage{bibentry}

\begin{document}

\maketitle

\begin{abstract}
Knowledge Tracing (KT) diagnoses students' concept mastery through continuous learning state monitoring in education.
Existing methods primarily focus on studying behavioral sequences based on ID or textual information.
While existing methods rely on ID-based sequences or shallow textual features, they often fail to capture (1) the hierarchical evolution of cognitive states and (2) individualized problem difficulty perception due to limited semantic modeling. 
Therefore, this paper proposes a \textbf{L}arge Language Model \textbf{H}yperbolic \textbf{A}ligned \textbf{K}nowledge \textbf{T}racing(\modelname).
First, the teacher agent deeply parses question semantics and explicitly constructs hierarchical dependencies of knowledge points; the student agent simulates learning behaviors to generate synthetic data. Then, contrastive learning is performed between synthetic and real data in hyperbolic space to reduce distribution differences in key features such as question difficulty and forgetting patterns. Finally, by optimizing hyperbolic curvature, we explicitly model the tree-like hierarchical structure of knowledge points, precisely characterizing differences in learning curve morphology for knowledge points at different levels.
Extensive experiments on four real-world educational datasets validate the effectiveness of our Large Language Model Hyperbolic Aligned Knowledge Tracing (\modelname) framework.
\end{abstract}

\begin{links}
\end{links}

\section{Introduction}


Knowledge Tracing (KT) ~\cite{abdelrahman2023knowledge}is a key technique in educational intelligence for tracking student knowledge states based on historical behaviors. It enables personalized instruction by assessing mastery levels~\cite{yin2023tracing,jones2004ubiquitous}. Existing methods are mainly divided into two categories: Sequence-centric approaches model student interactions as temporal processes~\cite{pandey2019self,piech2015deep}, capturing learning dynamics through patterns in response sequences. Graph-centric approaches, in contrast, encode structural relationships among concepts, leveraging graph neural networks to model dependencies.

\begin{figure}
    \centering
    \includegraphics[width=1\linewidth]{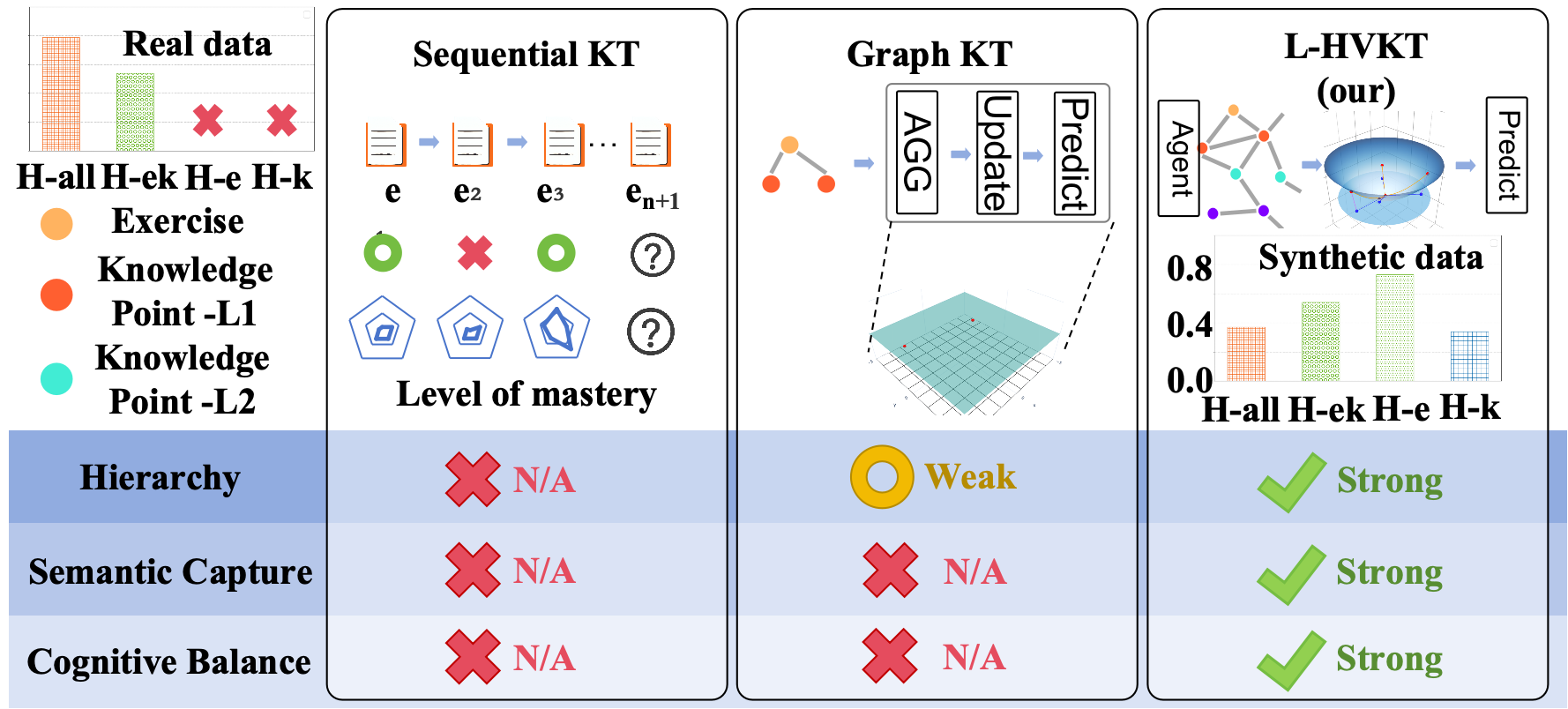}
    \caption{The bar chart labels H-all, H-ek, H-e, and H-k denote hyperbolicity measures for student-question-knowledge, question-knowledge, question-question, and knowledge-knowledge relationships, respectively. Lower values indicate stronger structural alignment with tree-like or hyperbolic space characteristics.}
    \label{fig:motivation}
\end{figure}

However, the existing methods rely on simple structured information and fail to fully exploit the rich semantics of questions~\cite{mislove2007measurement}\footnote{We interchangeably use the terms \textit{question} and \textit{exercise}  in this paper. }, making it difficult to capture the hierarchical dynamics of students' cognitive states. As shown in Figure \ref{fig:motivation}:
(1) Traditional methods model in Euclidean space~\cite{nickel2017poincare},  where the flat geometric structure cannot express the tree-like hierarchical properties of knowledge systems.
(2) The implicit topological relationships between knowledge concepts in the problem semantics have not been effectively captured~\cite{2015Deep}, resulting in the connections between the questions of related concepts not being well utilized.
(3) The degree of individual mastery distorts the difficulty of the practice~\cite{lee2022contrastive,chen2024modeling}. For instance, models trained on data from low-performing regions might wrongly classify moderately complex problems as high-difficulty ones, while high-performing students might label most problems as simple ones, which exposes the flaws of traditional difficulty assessment.

To bridge this gap, we attempt to utilize LLMs to extract knowledge point concepts~\cite{tan2021bidkt,tong2020exercise} from the questions to construct a knowledge graph, thereby modeling the potential correlations among different questions. It also enables LLMs to simulate human learning behaviors~\cite{gao2025agent4edu,piech2015deep,li2024bringing}, provide objective difficulty information of the questions, and construct a more accurate dynamic learning cognitive process. 

Despite this, there are still challenges in accurately simulating students' hierarchical cognition from simple to difficult:(1) Absence of hierarchical concept representation: The implicit hierarchical associations among knowledge points need to be modeled. For instance, the basic definition will gradually evolve into different inferences. Therefore, it is necessary to establish the hierarchical relationship of knowledge points to depict the learning trajectory of higher-order knowledge~\cite{fu2023hyperbolic,song2021hyperbolic}.
(2) Absence of hierarchical dynamic cognition: Although large models can simulate human behavior to assess the difficulty of questions, there are still different levels of understanding biases among individual students regarding the questions~\cite{park2023generative}. Therefore, it is necessary to align the general cognition of the difficulty level of agents with the dynamic understanding cognition of individual students.

In light of the above analysis, we propose a Large Language Model hyperbolic-aligned knowledge tracing method named \modelname{}. 
This method first utilizes the Teacher Agent to parse the semantic meaning of question texts, explicitly excavating the contained knowledge point concepts and their hierarchical dependencies, thereby resolving the absence of hierarchical representation for knowledge points. Through the Student Agent simulating individual learning behaviors, it generates synthetic data to supplement the cognitive process information missing in traditional data.
Then, we design a hyperbolic contrastive alignment mechanism:

Conducting contrastive learning between synthetic data and real student behavioral data in hyperbolic space to reduce distributional differences in key features such as question difficulty and forgetting patterns;
Finally, optimizing hyperbolic curvatures explicitly models the inherent tree-like hierarchy of knowledge points. This calibrates hierarchical-aware alignment between universal question difficulty perception and individual students' dynamic cognitive understanding, effectively resolving hierarchical dynamic cognitive deviation.
Consequently, the model precisely characterizes learning curve differences across knowledge levels and leverages hyperbolic space's hierarchical propagation to sensitively capture mastery leaps in higher-order knowledge points. The main contributions are as follows:

\begin{itemize}
\item  Our work is the first to propose an LLM dual-agent teacher-student collaborative framework, addressing the modeling blind spot of "student thinking paths" in traditional data.

\item  We innovatively introduce hyperbolic geometric space, achieving explicit modeling of knowledge point hierarchies through curvature optimization, and precisely characterizing learning curve morphology differences across knowledge levels.

\item  Based on aligned hierarchical data, we design a hyperbolic knowledge state tracker that leverages hierarchical propagation properties to sensitively capture mastery degree in higher-order knowledge points.
\end{itemize}

Experimental results demonstrate that the proposed \modelname{} possesses superior effectiveness and indeed enhances the capability of KT models in practical inference.

\section{Related Works}

\subsection{Knowlege Tracing}

Knowledge Tracing (KT) dynamically infers a student's knowledge state evolution by analyzing historical interactions and test data to predict future performance. Existing approaches bifurcate into two paradigms:
(1) Sequence Modeling-based KT: Representative works include Deep Knowledge Tracing (such as DKT ~\cite{piech2015deep},
ATKT~\cite{guo2021enhancing}, DIMKT~\cite{shen2022assessing}, etc.)  which pioneered the use of RNNs; the Dynamic Key-Value Memory Network (such as DKVMN~\cite{shen2022assessing}, SKVMN~\cite{zhang2017dynamic}, DGMN~\cite{abdelrahman2019knowledge}), 
 which explicitly updates knowledge point states via an external memory matrix; and Transformer architecture models (such as SAKT~\cite{pandey2019self} and SAINT~\cite{choi2020towards} that utilize self-attention mechanisms to capture long-range dependencies.
(2) Graph-based KT~\cite{tong2020structure,11231364}: Typical works include Graph-based Knowledge Tracing(GKT~\cite{nakagawa2019graph}), which explicitly constructs knowledge point adjacency graphs, and Structured Knowledge Tracing( SKT~\cite{tong2020structure}), which integrates dependency relationships from knowledge graphs.
These two categories of methods advance the refinement of knowledge state representation from the perspectives of temporal dynamics and structural relevance, respectively.
\subsection{Hyperbolic Machine Learning}

Hyperbolic geometric space excels in graph representation learning due to its hierarchical representation capability~\cite{fu2024hyperbolic,nickel2018learning,10891910}, making it ideal for tree-like structures such as knowledge graphs and user-item interaction graphs~\cite{icdm23DeepRicci}. It has been successfully applied in recommender systems,  for example with the LKGR model~\cite{chen2022modeling} capturing user-item hierarchies in Lorentz hyperbolic space. Significant progress has also occurred in knowledge graph embedding~\cite{www25RiemannGFM,nips25DeepMPNN}, including models like HHNE++~\cite{wang2019hyperbolic} that achieve low-distortion hierarchical representation of heterogeneous information networks~\cite{11180140,he2024detecting,icml24LSEnet,nips24MSN}. For recommender systems, hyperbolic space primarily optimizes user-item interactions~\cite{zhang2019star,wang2021hypersorec}, while for knowledge graphs it focuses on hierarchical dependencies between entities~\cite{perozzi2014deepwalk}. However, hyperbolic graph learning remains largely unexplored for intelligent education and knowledge tracing, particularly for hierarchical modeling of dynamic student-knowledge point interactions.
\section{Preliminary}

\subsection{Hyperbolic Geometry}
Hyperbolic space commonly refers to manifolds with constant negative curvature and is used for modeling complex networks~\cite{gao2022learning}. 
Among the common isometric models used to describe hyperbolic spaces, the hyperboloid model has recently been widely used in machine learning.
The hyperboloid model is an $n$-dimensional hyperbolic geometry as a manifold in the $(n+1)$-dimensional Minkowski space.
A hyperboloid manifold $\mathbb{H}^{n}_{\kappa} = \{x \in \mathbb{R}^{n+1} | \langle x, x \rangle_{\mathbb{H}} = 1/\kappa, \kappa < 0\}$ in $n$-dimensional space with curvature $\kappa$. 
$\langle \cdot, \cdot \rangle_{\mathbb{H}}$ is Lorentzian scalar product and $\langle x, y \rangle_{\mathbb{H}} := -x_0y_0 + x_1y_1 + \dots + x_ny_n$. The distance $d^{\mathbb{H}}_{\kappa}(x, y)$ on the hyperboloid model is defined as:
\begin{equation}
d^{\mathbb{H}}_{\kappa}(x, y) = \frac{1}{\sqrt{|\kappa|}} \operatorname{arccosh}(|\kappa|\langle x, y \rangle_{\mathbb{H}}).
\end{equation}

To ensure compatibility with standard machine learning methods, the logarithmic map is employed to project data into a local Euclidean tangent space ($\mathcal{T}_x\mathbb{H}^{n}_{\kappa}$) for computation~\cite{song2021hyperbolic}, and the exponential map is used to project the results back. The logarithmic map $\log^{\kappa}_x(\cdot)$ and exponential map $\exp^{\kappa}_x(\cdot)$ are defined as:
\begin{equation}
\log^{\kappa}_x(y) = d^{\mathbb{H}}_{\kappa}(x, y) \frac{y + |\kappa| \langle x, y \rangle_{\mathbb{H}}x}{\|y + |\kappa| \langle x, y \rangle_{\mathbb{H}}x\|_{\mathbb{H}}},
\end{equation}
\begin{equation}
\exp^{\kappa}_x(v) = \cosh\left(\sqrt{|\kappa|}\|v\|_{\mathbb{H}}\right)x + v\frac{\sinh\left(|\kappa|\|v\|_{\mathbb{H}}\right)}{\sqrt{|\kappa|}\|v\|_{\mathbb{H}}},
\end{equation}
where $\|v\|_{\mathbb{H}} = \sqrt{\langle v, v \rangle_{\mathbb{H}}}$ is the Lorentzian norm of $v$.

\subsection{Problem Statement}
Student Interaction Sequence: For each student $S$, we observe a chronologically ordered interaction sequence, denoted as $\mathcal{S}=\{{s}_{1},{s}_{2},\dots,s_m\}$. Interaction Content: Each interaction $s_j$ is a structured quadruple  $s_j = <q_j, \{c\}, r_j, t_j>$, where:
$q_j$ represents the question the student solved in the $j$-th interaction.
$\{c\}$ is the set of knowledge components associated with the question
$R_j \in \{0, 1\}$ is a binary response label. $R_j = 1$ indicates that student $S$ answered question $q_j$ correctly in the $j$-th interaction, while $R_j = 0$ indicates an incorrect answer.
$t_j$ represents the timestamp when the $j-th$ interaction occurred, recording the temporal information of the interaction.
Knowledge State and Embedding Space:$ k_t^c$ represents the latent knowledge state of student $S$ for knowledge concept $c$ at timestep $t$. 

\begin{figure}
    \centering
    \includegraphics[width=1\linewidth]{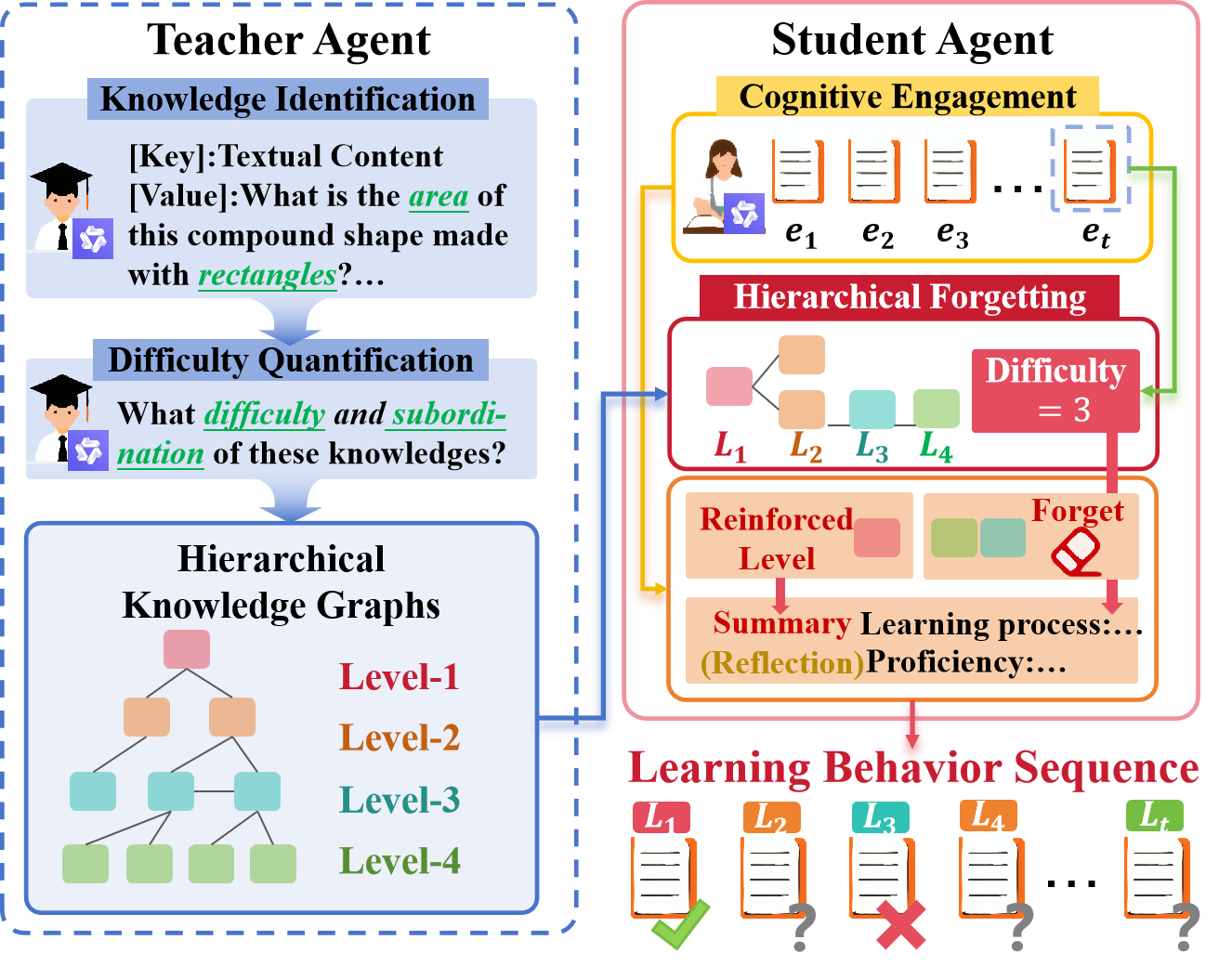}
    \caption{Illustration of LLM-based Teacher-student Behavior Modeling and Data Augmentation.}
    \label{fig:fra}
\end{figure}
\section{Method}

In this section, we propose our \modelname{} framework, aiming to assist knowledge tracing tasks through data generated by large models. It consists of three parts.
(1) We deploy two LLM-driven Agents,
Teacher Agent: Constructs hierarchical knowledge graphs by parsing exercise semantics and explicit knowledge dependencies.
Student Agent:Simulates personalized learning behaviors (e.g., engagement and forgetting) to generate synthetic interaction data.
(2) We introduce a relation-aware hyperbolic graph neural network 
, capturing hierarchical dependencies within knowledge systems by modeling hyperbolic embeddings with curvature optimization.
(3) We perform distribution calibration between Teacher-generated data and real-world student interactions via a hyperbolic contrastive alignment mechanism, integrating hierarchical states for knowledge tracing.

\subsection{LLM-Based Teacher-Student Behavior Augmentation}
Here, we employ two Agents:
(1) Teacher Agent: it extracts rich information from texts, constructing knowledge graphs.
(2) Student Agent: it captures learners' engagement states and behaviors to simulate their responses.
\subsubsection{Teacher Agent.}
The Teacher Agent $M_t$ processes problem images $q^{\textrm{image}}$ using specialized prompt templates to achieve three core outputs:
\begin{itemize}
    \item \textit{Hierarchical Knowledge Identification}: It parses images into semantic text $q^{text}_i = M_t(q^{\textrm{image}}_i)$ then extracts and classifies knowledge points into explicit hierarchy levels $L_j\in \{1,2,3,4\}$ (from basic definitions to comprehensive reasoning); 
    \item \textit{Structured Knowledge Graph Construction:} It builds a fine-grained hierarchical knowledge graph by establishing parent-child dependencies between knowledge points based on their levels, creating a tree-like pedagogical structure; 
    \item \textit{Exercise Difficulty Quantification:} It assigns objective difficulty scores to each exercise by analyzing the hierarchy levels of its associated knowledge points ${c^{L}_{i_1}, c^{L}_{i_2}, \cdots, c^{L}_{i_k}}$ , where higher-level knowledge combinations yield higher difficulty values.
\end{itemize}

This process provides: 1) Explicit knowledge hierarchy mapping, 2) Pedagogically structured knowledge graphs, and 3) Semantically-grounded exercise difficulty metrics for downstream components.
\subsubsection{Student Agent.}Utilizes the LLM's knowledge base and reasoning capabilities to simulate the problem-solving cognitive process of a real student. Given a student's historical interaction sequence ${l}_{j-1} = \{s_1, s_2, \dots, s_{j-1}\}$, where $s_i = \langle q_i, C_i, R_i \rangle$ contains the question $q_i$, associated knowledge points $C_i$, and response result $R_i \in \{0,1\}$ ,$R_i=1$ indicates correct,$R_i=0$ indicates error .
To simulate personalized learning behaviors, our Student Agent incorporates two specialized modules:
The Cognitive Engagement Module dynamically assesses the student's concentration level based on current exercise difficulty and mastery of related knowledge points.
The Hierarchical Forgetting Module models differential memory decay patterns according to knowledge point difficulty levels.
\begin{itemize}
    \item \textit{Cognitive engagement module}: A student's concentration during learning is influenced by problem difficulty $q_j$ and mastery of related knowledge points. 
    The learner's current engagement is defined as $\Gamma_j = \sigma(W_q \cdot [X_{q_i};X_{c_{ij}}
    ;t]+b_q)$. 
    If the interval $t$ for such problems is short, it indicates active learning.
    The current cognitive engagement state is evaluated based on embeddings of the 
    current problem $X_{q_i}$ and difficulty of related knowledge points $X_{c_{ij}}$.

    \item \textit{Hierarchical forgetting module}: We divide memory types into three levels based on problem difficulty. Basic knowledge points $L \in \{1,2\}$ cover core concepts, basic formulas, and simple rules.  Intermediate knowledge points $L \in \{3\}$ involve typical problem-solving methods, medium-complexity strategies, and combined applications. The forgetting curve has moderate slope. Difficult knowledge $L \in \{4\}$ includes complex problem decomposition. The forgetting curve is steepest with rapid decay. We model forgetting with $F_j= \exp(-\lambda \cdot L_{\mathrm{avg}} \cdot t_j)$, where $L_{\mathrm{avg}}$ is the average level of designed knowledge points.
\end{itemize}
Using the student's current engagement state and forgetting degree information for the current problem, we update the current knowledge state via LSTM~\cite{graves2012long}.
\begin{equation}
    \mathbf{h}_j^s = \mathbf{LSTM}([X_{q_j};\sum_{c\in {C_{ij}}}w_c{X}_c]\oplus(\Gamma_j\odot F_j \odot h_{j-1}^s))
\end{equation}
where $\mathbf{X}_{q_j}, \mathbf{X}_c$ are the question and knowledge point embeddings, respectively,$w_c$ is the weight of the knowledge point.

Based on the current exercise $e_j$ and historical state $h^{s}_{j-1}$, the student Agent predicts the student's response $A_j \in {0,1}$ and constructs/outputs the underlying reasoning path $\mathcal{P}_j = [c^{(1)} , \cdots ,c^{(k)} \rightarrow A_j]$, where $A_j$ is obtained through analyzing relevant knowledge points $c_j$ of the problem.

\subsection{Hyperbolic Encoding \& Alignment}
 
Since knowledge points are not isolated, we first construct a heterogeneous graph:
 $\mathcal{G}=(\mathcal{V},\mathcal{E})$, Then, our connecting exercise-knowledge point edges, and establishing dependency edges between knowledge points based on the knowledge graph.
$\mathcal{E}_{\mathrm{hie}}=\{(c_i,c_j)|L_{c_i}<L_{c_j}\}$ constructs directed hierarchical dependency chains (e.g., limit(level 2) → derivative(level 2) → differential equations(level 4)).

\subsubsection{Relational-aware Hyperbolic Graph Neural Network}

To model the complex hierarchical dependencies in both real $\kappa_{\text{real}}$ and synthetic $\kappa_{\text{syn}}$ graphs within hyperbolic space, we extend multi-head graph attention with curvature-specific processing. For each graph with its own curvature $\kappa \in \{\kappa_{\text{real}},\kappa_{\text{syn}}\}$, thereby defining two hyperbolic spaces $\mathbb{H}^{\kappa_{\text{1}}}$ and $\mathbb{H}^{\kappa_{\text{2}}}$ for real and synthetic graphs respectively, we project the initial Euclidean node embeddings $\mathbf{X}^{\mathbb{E}}$ into hyperbolic manifolds via the exponential map, yielding hyperbolic representations $\mathbf{H}_i^{\mathbb{H}_{\kappa}} = \exp_{\mathbf{0}}^{\kappa}(\mathbf{X}^{\mathbb{\mathbb{E}}})$. Finally, we perform relation-aware hyperbolic aggregation. Specifically, the representation for each node $i$ at layer $L$ is updated as follows:
\begin{equation}
\mathbf{h}_i^{{(L+1)}} = \exp_{\mathbf{0}}^{\kappa} \Bigg( \sigma \bigg( \sum_{j \in \mathcal{N}i} \alpha_{ij}^{(L)} \cdot \mathbf{W}^{(L)} \log_{\mathbf{0}}^{\kappa} \big( \mathbf{h}_j^{\mathbb{H}_{\kappa}^{(L)}} \big) \bigg) \Bigg)
\end{equation}
where $\alpha_{ij}^{(L)}$ is the attention score at layer $L$ with , and $\sigma$ is an activation function.

\subsubsection{Hyperbolic Constractive Alignment}

We use Hyperbolic Graph Neural Network (HGNN)~\cite{liu2019hyperbolic} to process $\mathcal{G}$, capturing the hierarchical nature of real and generated data. This embeds foundational knowledge points in the flatter central region, while distributing higher-order knowledge points in the higher-curvature peripheral regions, obtaining the $L$-th layer embeddings $h_v^{(L)}$,$h_u^{(L)}$, and apply contrastive learning to enhance their representational capability. For the exercise-concept side, exercises and knowledge concepts shared by both embedding spaces are labeled as positive samples, while other entities are treated as negative samples. Based on these positive and negative samples, the following contrastive loss is adopted:
\begin{equation}
    \mathcal{L}_{\mathrm{con}} = -\sum_{(u,v) \in \mathcal{P}} \log \frac{pos(h_u^{(L)},h_v^{(L)})}{pos(h_u^{(L)},h_v^{(L)}) + neg(h_u)},
\end{equation}
where $\mathcal{P}$ is positive pairs, $ sim(h_u^{(L)}, h_v^{(L)}) $ denotes the cosine similarity between embeddings of $ h_u^{(L)} $ and $ h_v^{(L)} $, $ \tau $ is a temperature coefficient.
\begin{figure*}
    \centering
    \includegraphics[width=0.96\linewidth]{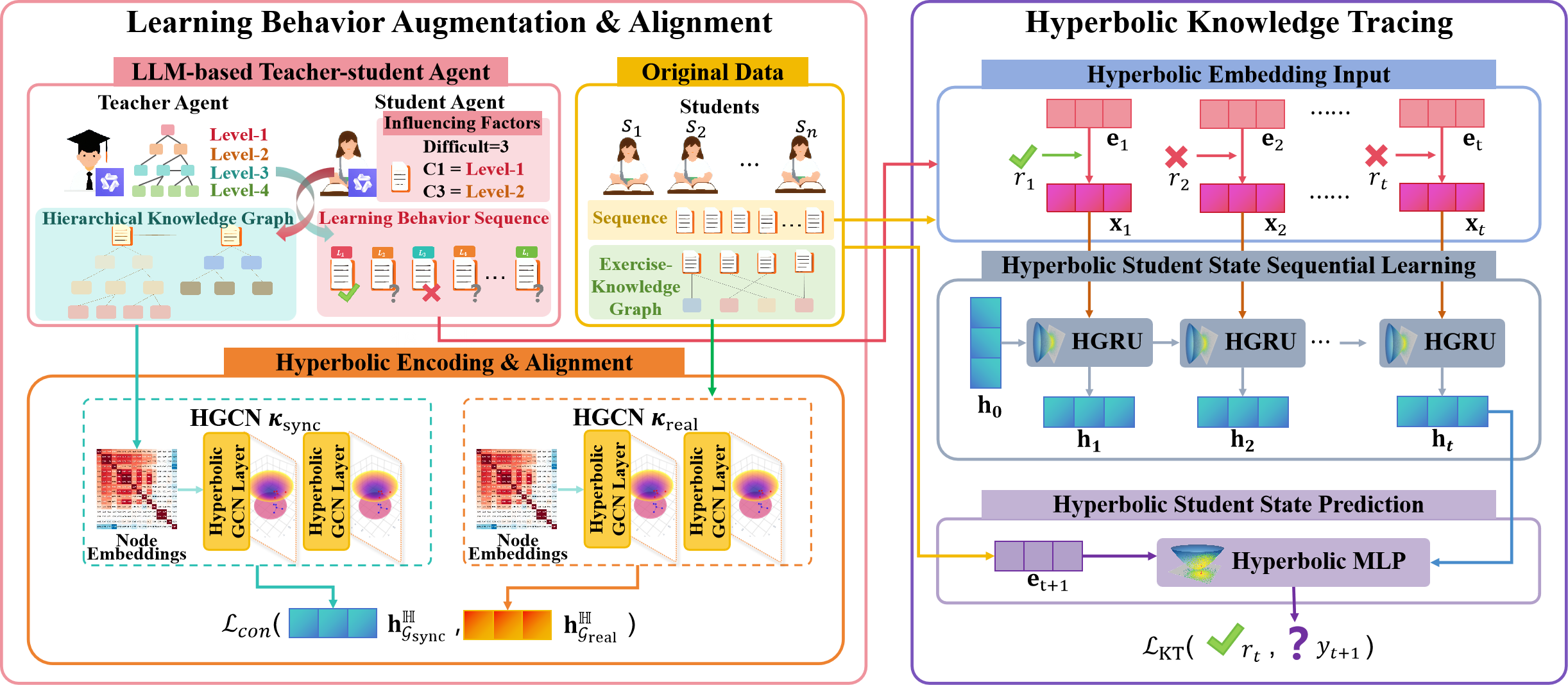}
    \caption{Illustration of the architecture of L-HAKT.}
    \label{fig:Hyperbolic}
\end{figure*}

\subsection{Hyperbolic Knowledge Tracing}
\subsubsection{Sequence Propagation}
We obtain hyperbolic embeddings for exercises and associated knowledge points $X_{qc}^{\mathbb{H}}$ 
  from our relation-aware hyperbolic graph neural network. To incorporate the correctness rate $R_t$ into these embeddings while maintaining computational efficiency, we first project $X_{qc}^{\mathbb{H}}$to the tangent space at the origin via the logarithmic map  $X_{qc}^{\mathbb{E}} = \log_0^\kappa(X_{qc}^{\mathbb{H}})$ , then perform the update 
  $X_i=\phi (X_{qc}^{\mathbb{H}}\oplus R_t)$in Euclidean space. For sequential processing, we similarly map the previous hyperbolic hidden state $H_{t-1}^{\mathbb{H}}$ to its Euclidean counterpart $H_{t-1}^{\mathbb{E}} = \log_0^\kappa(H_{t-1}^{\mathbb{H}})$.We then construct joint representations by integrating question  difficulty
$q^{\textrm{diff}}$ and the average knowledge point level $\overline{c^L}$,applying hierarchical differentiation through $D^{t} = \phi(W_d\cdot q^{\textrm{diff}}+W_L\cdot \overline{c^L})$ where higher-level problems receive reinforced processing and basic problems undergo standard handling. This dual-space approach leverages hyperbolic geometry's hierarchical representation capabilities while maintaining computational tractability through tangent space operations~\cite{yang2022hyperbolic}.

\begin{equation}
    {H}_t^{\mathbb{E}} = HGRU(X^{\mathbb{E}}_t,{H}_{t-1}^{\mathbb{E}},D_t)
\end{equation}
As the GRU is built upon the tangent space, logarithmic maps are needed . Then, we feed the states into the GRU and map the hidden state back to hyperbolic space $H^{\mathbb{H}}_t=exp_0^\kappa({H}_t^{\mathbb{E}})$. As we can see, the final $H^{\mathbb{H}}$ is capable of integrating the knowledge point hierarchy structure, content, and question-solving information.

\subsubsection{Model Training}

The KT loss $\mathcal{L}_{kt}$ is defined as the binary cross-entropy loss between the prediction $y_t$ and the correct answer $\hat{y}$, calculated as follows:
\begin{equation}
     \mathcal{L}_{\mathrm{KT}} = -\sum_{i=1}\left(y_t log\hat{y} +\left(1-y_t\right)log\left(1-\hat{y}\right) \right)
\end{equation}

\begin{equation}
    \mathcal{L}_{\mathrm{total}} = \mathcal{L}_{\mathrm{KT}} + \alpha \mathcal{L}_{\mathrm{con}} 
\end{equation}
where $\alpha$ is the hyperparameter controlling $\mathcal{L}_{con}$.

\section{Experiment}
\begin{table*}[htbp]
\centering
\small
\setlength{\tabcolsep}{1.5mm}
\begin{tabular}{c|cc|cc|cc|cc}
\toprule
\multirow{2}{*}{Model}&\multicolumn{2}{c|}{ASSIST09}&\multicolumn{2}{c|}{ASSIST12}&\multicolumn{2}{c|}{EdNet}&\multicolumn{2}{c}{Eedi}
\\
\cmidrule(lr){2-3}\cmidrule(lr){4-5}\cmidrule(lr){6-7}\cmidrule(lr){8-9}
&AUC(\%)&ACC(\%)&AUC(\%)&ACC(\%)&AUC(\%)&ACC(\%)&AUC(\%)&ACC(\%)
\\
\midrule
DKT&75.97${\pm0.25}$&73.01${\pm0.21}$&72.90${\pm0.07}$&73.27${\pm0.16}$&70.10${\pm0.37}$&71.11${\pm0.42}$&76.01${\pm0.03}$&71.64${\pm3.81}$\\
DKT+&77.21${\pm0.03}$&73.32${\pm0.21}$&73.64${\pm0.01}$&73.42${\pm0.01}$&70.20${\pm0.06}$&66.21${\pm1.67}$&74.32${\pm0.01}$&70.79${\pm0.01}$\\
DKT+forget&77.32${\pm0.00}$&72.13${\pm0.16}$&73.75${\pm0.03}$&73.12${\pm0.15}$&70.46${\pm0.03}$&71.12${\pm0.17}$&76.13${\pm0.05}$&71.66${\pm0.14}$\\
DIMKT&72.01${\pm0.36}$&73.68${\pm0.16}$&76.23${\pm0.22}$&74.98${\pm0.02}$&77.00${\pm0.19}$&72.93${\pm0.02}$&79.30${\pm0.02}$&73.03${\pm0.90}$\\
DKVMN&76.10${\pm0.11}$&72.34${\pm0.07}$&72.12${\pm0.21}$&73.32${\pm0.23}$&69.32${\pm0.14}$&70.95${\pm0.02}$&76.01${\pm0.29}$&71.62${\pm0.09}$\\
Deep-IRT&76.21${\pm0.11}$&72.10${\pm2.17}$&72.32${\pm0.36}$&73.64${\pm0.00}$&69.73${\pm0.08}$&71.22${\pm0.03}$&76.00${\pm0.37}$&71.73${\pm0.20}$\\
ATKT&76.53${\pm0.90}$&73.33${\pm0.01}$&73.24${\pm0.05}$&72.98${\pm0.02}$&70.23${\pm0.06}$&72.10${\pm0.03}$&76.21${\pm0.22}$&72.01${\pm0.36}$\\
AT-DKT&76.89${\pm0.02}$&73.05${\pm0.04}$&72.74${\pm0.07}$&72.90${\pm0.02}$&70.11${\pm0.04}$&71.03${\pm0.21}$&75.94${\pm0.04}$&72.00${\pm0.11}$\\
AKT&78.23${\pm0.05}$&74.45${\pm0.30}$&78.21${\pm0.11}$&75.23${\pm0.01}$&76.78${\pm0.08}$&73.32${\pm0.10}$&78.84${\pm0.02}$&73.01${\pm0.16}$\\
SAKT&75.33${\pm0.03}$&71.77${\pm0.08}$&73.03${\pm0.58}$&73.40${\pm0.18}$&69.27${\pm0.05}$&71.07${\pm0.21}$&75.73${\pm0.18}$&71.23${\pm0.09}$\\
SAINT&75.00${\pm0.03}$&71.21${\pm4.85}$&75.11${\pm0.09}$&74.01${\pm0.03}$&76.04${\pm0.24}$&73.45${\pm0.12}$&78.32${\pm0.08}$&72.93${\pm0.14}$\\
CL4KT&76.26${\pm0.14}$&72.75${\pm0.30}$&72.36${\pm0.29}$&73.31${\pm3.11}$&69.65${\pm0.06}$&71.18${\pm0.01}$&75.83${\pm0.03}$&71.47${\pm0.07}$\\
GKT&76.32${\pm0.21}$&72.44${\pm0.14}$&72.39${\pm0.04}$&73.29${\pm0.06}$&69.21${\pm0.14}$&71.01${\pm0.01}$&76.02${\pm2.17}$&71.68${\pm0.05}$\\
GIKT &77.33${\pm0.02}$&72.69${\pm0.24}$&76.32${\pm0.05}$&75.11${\pm0.02}$&76.02${\pm0.21}$&73.11${\pm0.10}$&\underline{79.68${\pm0.08}$}&73.02${\pm0.20}$\\
simpleKT&77.12${\pm0.42}$&73.56${\pm0.13}$&77.21${\pm0.07}$&75.49${\pm0.14}$&75.11${\pm0.03}$&73.57${\pm0.03}$&78.22${\pm0.33}$&\underline{73.12${\pm0.03}$}\\
MIKT&\underline{79.38${\pm0.03}$}&\underline{74.54${\pm0.01}$}&\underline{78.65${\pm0.03}$}&\underline{76.52${\pm0.07}$}&\underline{77.10${\pm0.10}$}&\underline{74.22${\pm0.30}$}&79.59${\pm0.07}$&72.67${\pm0.90}$\\
\midrule
our&\textbf{80.22${\pm0.30}$}&\textbf{75.10${\pm0.04}$}&\textbf{80.27${\pm0.40}$}&\textbf{77.10${\pm0.14}$}&\textbf{78.23${\pm0.05}$}&\textbf{75.21${\pm0.01}$}&\textbf{80.29${\pm0.02}$}&\textbf{73.49${\pm0.03}$}
\\
\bottomrule
\end{tabular}

\caption{Overall AUC and ACC performance of L-HAKT and all baselines. ACC and AUC should be as large as possible, indicating better model performance; \textbf{Bold}:best; \underline{Underline}: runner-up.}
\label{tab:main}
\end{table*}

\begin{table}
  \centering
    \small
      \setlength{\tabcolsep}{1mm}
    \begin{tabular}{lcccc}
    \toprule
    & \textbf{ASSIST09} & \textbf{ASSIST12} & \textbf{EdNet} & \textbf{Eedi}  \\
    \midrule
    \# Students &  4160& 5000 & 5000 & 5000 \\
    \# Question & 15643 & 36054 & 11700 & 26702\\
    \# Concept & 167 & 242 & 1830 & 1050 \\
    \# Interaction & 206631 & 713123 & 1147423 & 586234 \\
   
    \bottomrule
    \end{tabular}
  \caption{Summary of dataset statistics.}
  \label{tab:dataset_statistics}
\end{table}

\begin{table*}[ht]
\centering
\small
\begin{tabular}{c  cc cc  cc cc}
\toprule
\multirow{2}{*}{Model} 
& \multicolumn{4}{c}{ASSIST09} 
& \multicolumn{4}{c}{ASSIST12} \\
\cmidrule(lr){2-5} \cmidrule(lr){6-9}
& ACC(\%) & $\Delta$(\%) & AUC(\%) & $\Delta$(\%) 
& ACC(\%) & $\Delta$(\%) & AUC(\%) & $\Delta$(\%) \\
\midrule
GKT & 76.32 & - & 72.44 & - & 72.39 & - & 73.29 & - \\
L-HVKT(w/o hyp) & \underline{77.55} & \underline{$\uparrow$1.61} & \underline{73.78} & \underline{$\uparrow$1.85} & \underline{78.01} & \underline{$\uparrow$7.76} & \underline{76.84} & \underline{$\uparrow$4.84} \\
L-HVKT(w/o con) & 76.98 & $\uparrow$0.86 & 73.21 & $\uparrow$1.06 & 76.32 & $\uparrow$5.43 & 75.34 & $\uparrow$2.80 \\
L-HVKT & \textbf{80.22} & \textbf{$\uparrow$5.11} & \textbf{75.10} & \textbf{$\uparrow$3.67} & \textbf{80.27} & \textbf{$\uparrow$10.89} & \textbf{77.10} & \textbf{$\uparrow$5.20} \\
\bottomrule
\end{tabular}

\begin{tabular}{c  cc cc  cc cc}
\multirow{2}{*}{Model} 
& \multicolumn{4}{c}{EdNet} 
& \multicolumn{4}{c}{Eedi} \\
\cmidrule(lr){2-5} \cmidrule(lr){6-9}
& ACC(\%) & $\Delta$(\%) & AUC(\%) & $\Delta$(\%) 
& ACC(\%) & $\Delta$(\%) & AUC(\%) & $\Delta$(\%) \\
\midrule
GKT & 69.21 & - & 71.01 & - & 76.02 & - & 71.68 & - \\
L-HVKT(w/o hyp) & \underline{76.54} & \underline{$\uparrow$10.59} & \underline{74.53} & \underline{$\uparrow$4.96} & \underline{79.02} & \underline{$\uparrow$3.95} & \underline{72.48} & \underline{$\uparrow$1.12} \\
L-HVKT(w/o con) & 75.51 & $\uparrow$9.10 & 72.67 & $\uparrow$2.34 & 78.19 & $\uparrow$2.85 & 71.97 & $\uparrow$0.40 \\
L-HVKT & \textbf{78.23} & \textbf{$\uparrow$13.03} & \textbf{75.21} & \textbf{$\uparrow$5.91} & \textbf{80.29} & \textbf{$\uparrow$5.59} & \textbf{73.49} & \textbf{$\uparrow$2.53} \\
\bottomrule
\end{tabular}
\caption{The ACC and AUC improvements (\%) results of Ablation Study. (\textbf{Bold}:best; \underline{Underline}:runner-up.)}
\label{tab:ablation_ednet_eedi}
\end{table*}
\subsection{Dataset Description}
We conducted experiments on four public education datasets: ASSIST2009 \cite{feng2009addressing}, ASSIST2012  \cite{feng2009addressing}, EdNet \cite{choi2020ednet}, and Eedi\cite{wang2020neural}. These datasets comprehensively record student-question-knowledge point interaction triples, with Eedi additionally containing image information corresponding to questions. To address Eedi's multimodal characteristics, we utilized a vision-Large Language Model (e.g., Qwen-2.5VL) to parse question image content and simulate student problem-solving processes: first converting images to textual descriptions, then combining question text prompts to simulate answer behaviors of students at different cognitive levels via LLM. Detailed dataset statistics are shown in Table \ref{tab:dataset_statistics}, which specifically annotates the volume of simulated interaction data newly added to Eedi after LLM enhancement.
\subsection{Experimental Setup}
Similar to (~\cite{sun2024interpretable,liu2023simplekt}), we randomly partitioned the dataset into two subsets, with 80\% used for training and 20\% used for testing.
We evaluated model performance using Accuracy (Acc) and AUC. Accuracy measures classification correctness, while AUC assesses the model’s ability to distinguish between different classes.
All models were trained and tested on a single Nvidia A100 40GB GPU. The vector dimension $d$ was set to match the number of knowledge concepts. The representation learning module consists of 2 layers. A hyperparameter search was conducted for $\alpha$ over the range $[0.0001, 0.001, 0.01, 0.1, 1]$. 

\subsection{Baseline Approaches}
To validate the effectiveness of our  \modelname{}, we compare it against various baseline methods from two categories.

\textbf{Sequence Models:} 
This category encompasses temporal knowledge tracing approaches including recurrent-based methods like DKT~\cite{piech2015deep} using LSTM for knowledge state tracking alongside its enhanced variants DKT+\cite{yeung2018addressing} addressing inconsistent states, DKT+forgetting\cite{nagatani2019augmenting} incorporating forgetting behavior. Attention mechanisms are represented by ATKT~\cite{guo2021enhancing} employing adversarial training and AT-DKT utilizing auxiliary tasks.DIMKT~\cite{shen2022assessing} examines the impact of problem difficulty on students' knowledge levels. Memory network approaches feature DKVMN~\cite{zhang2017dynamic} with dynamic key-value memory and Deep-IRT integrating item response theory. Transformer architectures include SAKT~\cite{pandey2019self} capturing relationships through self-attention, SAINT~\cite{choi2020towards} implementing full Transformer modeling, AKT~\cite{ghosh2020context} simulating forgetting through context attention. CL4KT~\cite{lee2022contrastive} addressing interaction sparsity, and DTransformer tracking stable states. simpleKT~\cite{liu2023simplekt} is based on the simplified AKT model structure, achieving simplicity without sacrificing performance.MIKT~\cite{sun2024interpretable} tracks the students' state of domain knowledge and conceptual knowledge.

\textbf{Graph Models:}  This category focuses on explicit knowledge topology modeling through graph structures, featuring GKT~\cite{nakagawa2019graph} which propagates conceptual knowledge states via graph neural networks alongside GIKT~\cite{yang2020gikt} that enhances question representations using graph convolutional networks to aggregate question-concept relationships. These methods directly leverage graph connectivity to capture dependencies between knowledge points.

\subsection{Performance Comparison}
As shown in Table \ref{tab:main},  \modelname{} significantly outperforms both sequence models and graph model baselines across four real-world datasets. This advantage verifies that hyperbolic geometric space can effectively capture the latent hierarchical structure among knowledge points, adaptively distinguishing learning dynamics differences between fundamental and advanced knowledge points through hyperbolic curvature. Simultaneously, the data generated by large language models substantially enhances the performance of graph models in knowledge tracing tasks, highlighting the critical role of synthetic data in strengthening model generalization capabilities.
\begin{figure*}
    \centering
    \subfigure[Comparison of the performance with different knowledge graphs.]{
    \includegraphics[height=0.152\textheight]{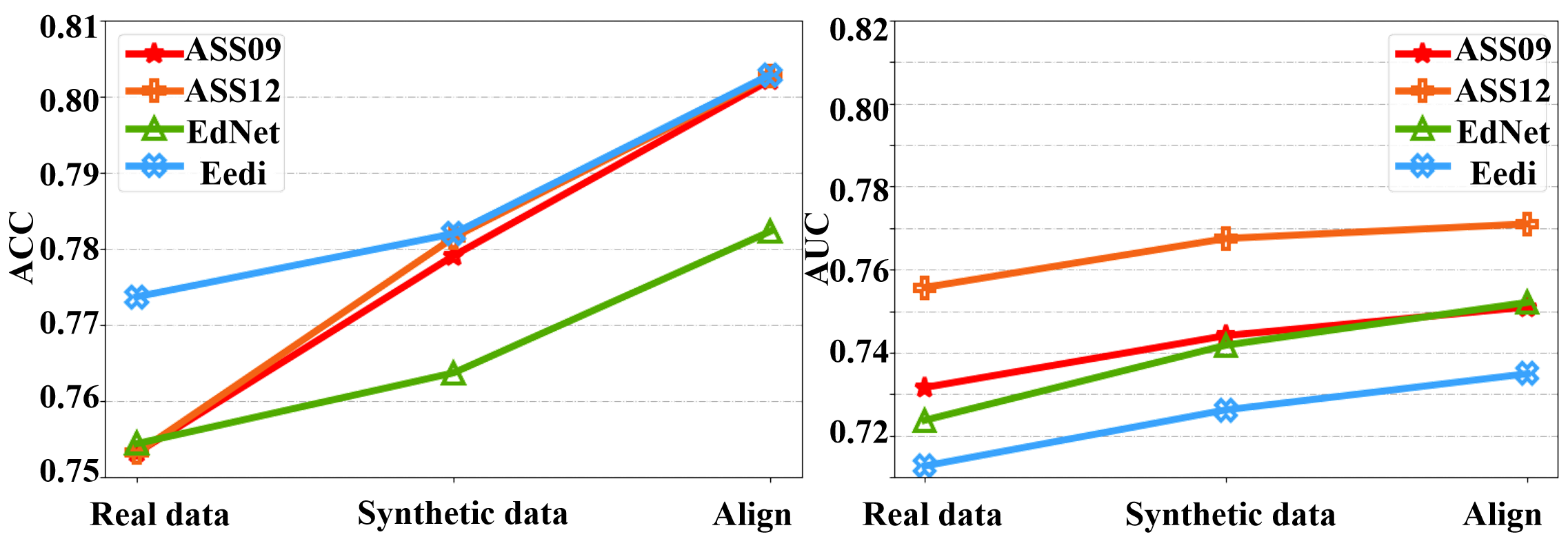}
    }
    \subfigure[Validation of knowledge generation of teacher agent.]{
    \includegraphics[height=0.152\textheight]{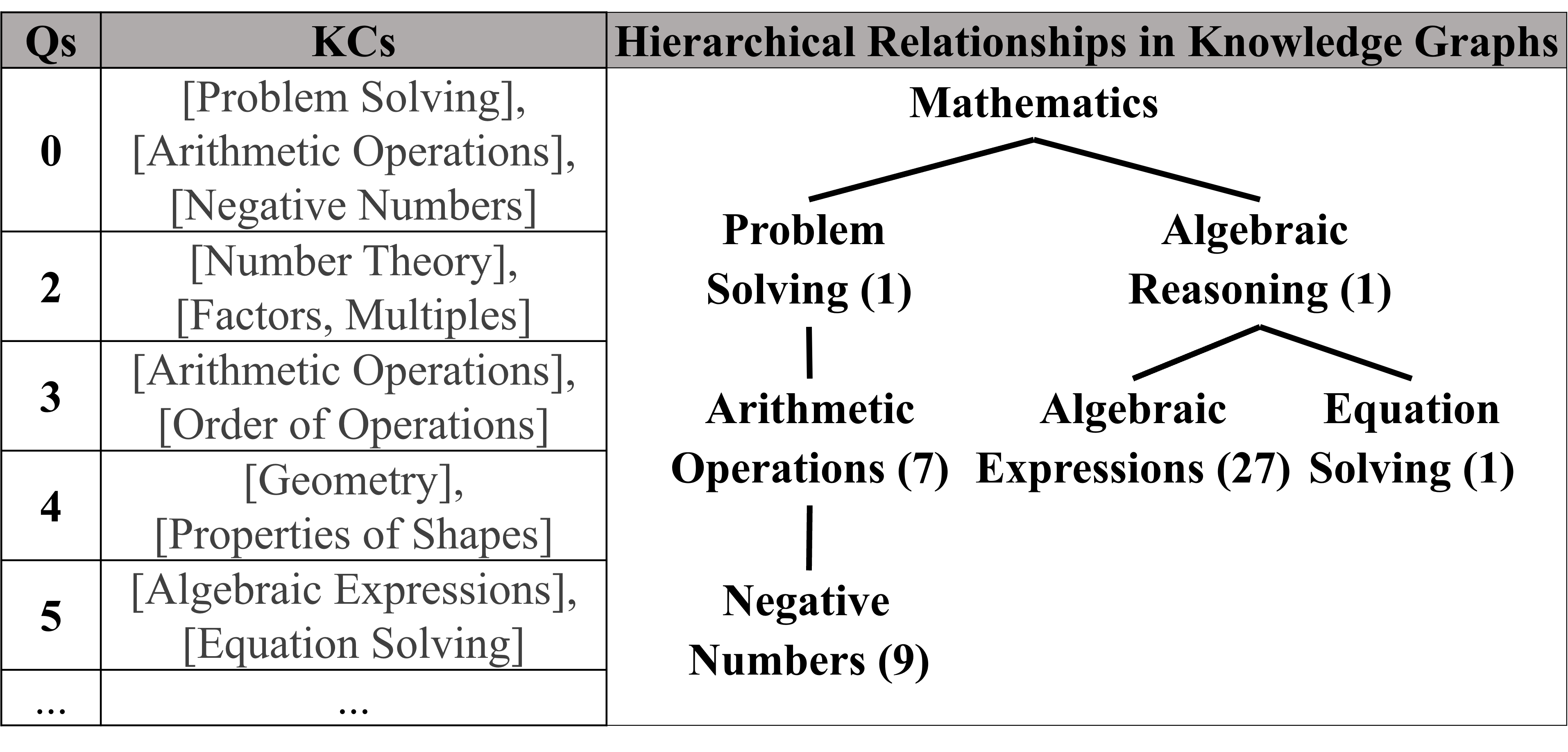}
    }
    \caption{Validation of Knowledge Graph Effectiveness. (a)
Compare the predictive performance of ACC and AUC in four datasets using real data, adding synthetic data, and comparing alignment.(b)Teacher Agent generates a knowledge graph by parsing rich text information.Where (.) denotes the number of related concepts.}
    \label{fig:align}
\end{figure*}
\subsection{Ablation Study}
We conducted ablation studies to evaluate \modelname{} two core mechanisms: hyperbolic representation learning and multi-level contrastive alignment, using Graph-based Knowledge Tracing (GKT) as the backbone network. Experimental results (Table 3) compare three configurations: the full L-HAKT, \modelname{} (w/o con) without contrastive alignment, and \modelname{} (w/o hyp) without hyperbolic representation. Hyperbolic representation learning significantly enhanced hierarchical structure capture, where its adaptive curvature mechanism effectively distinguished learning dynamics between fundamental and advanced knowledge points—particularly crucial for modeling mutation characteristics in complex skills. The performance degradation in \modelname{} (w/o hyp) confirms Euclidean space's limitation in leveraging rich hierarchical relationships. Meanwhile, multi-level contrastive alignment played a vital role in correcting LLM-generated data distribution bias; its removal in \modelname{} (w/o con) demonstrates that synthetic data alone merely provides data augmentation benefits, whereas proper alignment enables authentic distribution calibration and cognitive plausibility enhancement. This module substantially increased sensitivity to mutation states in advanced knowledge points. Collectively, the geometric embedding of knowledge hierarchies in hyperbolic space combined with precise distribution alignment through contrastive learning enables \modelname{} consistent outperformance of its variants across all four datasets.

\subsection{Validation of Knowledge Graph Effectiveness}
To rigorously validate the effectiveness of our hierarchical knowledge graph, Figure ~\ref{fig:align}.a compares three methodological configurations: the baseline restricted to original question-knowledge point connections, the augmented approach incorporating Student Agent synthetic data, and the full framework integrating Teacher Agent knowledge graphs with hyperbolic contrastive alignment. The results demonstrate that LLM-synthetic data substantially enriches sparse learning trajectories by simulating diverse cognitive pathways. 
Simultaneously, the contrastive alignment mechanism effectively bridges distributional discrepancies between synthetic and real behavioral patterns while preserving pedagogical hierarchies through geometric constraints in hyperbolic embedding space.
Complementing these quantitative insights, Figure ~\ref{fig:align}.b visually substantiates our approach through a representative knowledge graph segment. 
This topology exhibits pedagogically coherent hierarchical dependencies.
Collectively, these outcomes confirm our framework successfully resolves two fundamental limitations of traditional knowledge tracing: the fragmentation of knowledge structures through hierarchical graph construction, and the distortion of difficulty perception via geometrically grounded representation.

\section{Conclusion}

We propose \textbf{L}arge Language Model \textbf{H}yperbolic \textbf{A}ligned \textbf{K}nowledge \textbf{T}racing(\modelname), a knowledge tracing framework integrating large language models and hyperbolic geometry to systematically model how hierarchical knowledge structures distinctly influence learning dynamics—progressive mastery of basic concepts versus abrupt evolution of advanced knowledge points. Our dual-agent mechanism employs a Teacher Agent to build logically dependent knowledge graphs and a Student Agent to simulate cognition-driven problem-solving behaviors, generating synthetic data rich in analytical pathways. Through hyperbolic contrastive alignment, we simultaneously calibrate distributions between synthetic and real data while geometrically embedding knowledge hierarchies in curvature-adaptive hyperbolic space. Experiments across four real-world education datasets validate the framework's effectiveness, with hyperbolic embeddings visually demonstrating correlations between knowledge hierarchies and learning curve morphologies. Future work will explore hyperbolic learning generation for instructional optimization and leverage combined model-space  for fine-grained diagnostics.

\section{Acknowledgments}
Corresponding author is Dongran Yu. This paper is supported by the National Natural Science Foundation of China (No.62462007), The Basic Ability Enhancement Program for Young and Middle-aged Teachers of Guangxi (No.2024KY0073),  the Research Fund of Guangxi Key Lab of Multi-source Information Mining \& Security (MIMS24-12), and 2024 Guangxi Normal University Joint Training Project of the National Natural Science Foundation (No.2024PY029). We sincerely thank all the authors for their hard work and contributions.

\bibliography{aaai2026}

\end{document}